\documentclass[sigconf,table]{acmart}

\usepackage{amsmath}
\usepackage{graphicx} 
\usepackage[ruled,vlined]{algorithm2e}
\usepackage{paralist}
\usepackage{tikz}
\usetikzlibrary{positioning, arrows.meta}
\usepackage{multirow}

\usepackage{listings}
\usepackage{xcolor}
\definecolor{top3}{RGB}{255, 223, 186}  
\definecolor{top1}{RGB}{204, 255, 229}  
\definecolor{top2}{RGB}{204, 229, 255}  

\AtBeginDocument{%
  }

\begin{document}

\title{Efficient Graph Understanding with LLMs via Structured Context Injection}


\author{Govind Waghmare}
\affiliation{%
  \institution{Mastercard}
  \country{India}
  }
\email{govind.waghmare@mastercard.com}

\author{Sumedh BG}
\affiliation{%
  \institution{Mastercard}
  \country{India}
  }
\email{sumedh.sumedh@mastercard.com}

\author{Sonia Gupta}
\affiliation{%
  \institution{Mastercard}
  \country{India}
  }
\email{sonia.gupta@mastercard.com}

\author{Srikanta Bedathur}
\affiliation{%
  \institution{Indian Institute of Technology, Delhi}
  \country{India}
  }
\email{srikanta@cse.iitd.ac.in}








\begin{abstract}

Large Language Models (LLMs) have shown strong capabilities in solving problems across domains, including graph-related tasks traditionally addressed by symbolic or algorithmic methods. In this work, we present a framework for structured context injection, where task-specific information is systematically embedded in the input to guide LLMs in solving a wide range of graph problems. Our method does not require fine-tuning of LLMs, making it cost-efficient and lightweight. We observe that certain graph reasoning tasks remain challenging for LLMs unless they are mapped to conceptually grounded representations. However, achieving such mappings through fine-tuning or repeated multi-step querying can be expensive and inefficient. Our approach offers a practical alternative by injecting structured context directly into the input, enabling the LLM to implicitly align the task with grounded conceptual spaces. We evaluate the approach on multiple graph tasks using both lightweight and large models, highlighting the trade-offs between accuracy and computational cost. The results demonstrate consistent performance improvements, showing that structured input context can rival or surpass more complex approaches. Our findings underscore the value of structured context injection as an effective and scalable strategy for graph understanding with LLMs.

\end{abstract}





\maketitle

\section{Introduction}

LLMs have demonstrated remarkable potential in solving problems across a wide range of domains beyond the natural language tasks they were originally designed for \cite{Devlin2019BERTPO, vaswani2017attention, achiam2023gpt, reid2024gemini, touvron2023llama}. They have recently been applied to various graph-based tasks traditionally solved by algorithmic methods, such as reachability, connectivity, topological sorting, bipartite graph matching, shortest path computation, and many more \cite{wang2024canNLGraph, fatemi2024talk, perozzi2024let, ye2023languageInstructGLM, yu2024thought, 10.1145/3655103.3655110}. These promising results open up interesting possibilities for using LLMs in fields like network optimization, computational biology, and even dynamic system modeling. Unlike traditional graph algorithms, LLMs can interpret graph problems within broader, real-world contexts, offering a flexible approach to reasoning through complex graph structures. By incorporating task-specific context and web-based knowledge, LLMs can add valuable insights, improving the interpretation and solution of tasks like connectivity, shortest paths, and graph traversal. This adaptability makes LLMs a promising tool for enhancing problem-solving in graph-based applications.

Despite recent progress, several key challenges persist in using LLMs for solving graph-related problems:

\begin{compactenum}
\item As graph structures grow in complexity, even advanced models like GPT-4 show limitations in accurately reasoning over them. This suggests a need for better alignment with grounded conceptual representations to enable effective graph understanding \cite{Lin2024GraphenhancedLLPLAG, wang2024canNLGraph}.

\item High computational costs, both in terms of large model size and the need for multi-step querying or fine-tuning, can make LLM-based solutions inefficient, especially for large-scale graph data. The trade-off between performance and resource use calls for more cost-effective alternatives.

\item Existing methods often rely on generic prompt templates that fail to incorporate task-specific context. Such unstructured input limits the model’s ability to adapt to the nuances of different graph problems, highlighting the need for more structured and targeted context injection.
\end{compactenum}

Addressing these challenges in a  systematic manner is crucial for realizing the full potential of LLMs in reasoning with graphs and achieving practical, reliable applications in real-world scenarios.

\subsection{Our Proposal}

In response to the challenges identified, we propose a structured context injection framework that enhances LLM performance on graph-related tasks without requiring fine-tuning or expensive multi-step querying. Unlike previous approaches that rely on generic or static templates, our method introduces task-specific constraints and graph-aware contextual cues directly into the model input. This targeted inclusion of structured knowledge improves the LLM's ability to reason over complex graph structures while maintaining cost-efficiency.

Our framework is model-agnostic, supporting both large and lightweight LLMs, and is applicable across a range of graph problems expressed in natural language. Crucially, this approach avoids the need for grounding via fine-tuning or large prompt chains, enabling practical deployment on real-world datasets with minimal overhead. To demonstrate the impact of our approach, we carry out practical and relevant evaluations across diverse graph tasks and datasets. We also compare the performance and efficiency of different LLM sizes to highlight the effectiveness of our structure-aware context injection in enabling robust graph reasoning.

\noindent \textbf{Key Contributions:}
\begin{itemize}
\item \textbf{Structured Context Injection:} This work proposes a new strategy for encoding task-specific and structural cues within LLM inputs to enable more accurate and computationally efficient graph reasoning.

\item \textbf{Model-Agnostic Design:} Our framework avoids model fine-tuning and is compatible with any pre-trained LLM, enabling broad applicability to graph tasks such as connectivity checking, topological sorting, and shortest path.

\item \textbf{Efficiency vs. Performance Trade-off:} We conduct a comparative analysis between a compact and a large-scale LLM, showing that with the right context, smaller models can achieve comparable results at a fraction of the cost.

\end{itemize}

\noindent Through these contributions, we offer a practical, scalable solution that bridges the gap between graph structure and LLM reasoning, delivering improvements without compromising efficiency.

\section{Related Work}

Recent progress in graph reasoning has leveraged prompting strategies and benchmarks tailored to graph tasks. For instance, InstructGLM \cite{ye2023languageInstructGLM} developed instruction-tuned LLMs for node classification, while NLGraphs \cite{wang2024canNLGraph} introduced a benchmark covering eight graph problems using algorithmic prompting. Other works explore LLMs’ ability to maintain structured models for reasoning \cite{adhikari, Ammanabrolu}, perform multi-hop question answering \cite{Creswell, yu2023crepe}, and generate diverse graph types such as event and explanation graphs \cite{tandon-etal-2019-wiqa, saha-etal-2021-explagraphs}. Reasoning frameworks like Tree-of-Thoughts (ToT) \cite{yao2023tree}, Graph-of-Thoughts (GoT) \cite{besta2024graph}, Buffer-of-Thoughts (BoT) \cite{yang2024buffer} and Monte Carlo Tree Search (MCTS) \cite{zhang2024accessingMCTS} organize intermediate reasoning steps but often incur high inference costs.

Despite these advances, existing approaches face limitations in generalizability across diverse graph tasks and often rely on fine-tuning or architectural modifications \cite{clark_finetune, dwivedi2021generalization}, which increase complexity and cost. Moreover, the trade-offs between performance and inference efficiency remain underexplored. Our work addresses these gaps by proposing a modular and cost-effective structured context injection method that improves LLM graph reasoning without fine-tuning or model changes, enabling fair comparisons across different LLMs.

\section{Proposed Approach}

\subsection{Injecting Structured Task Context}
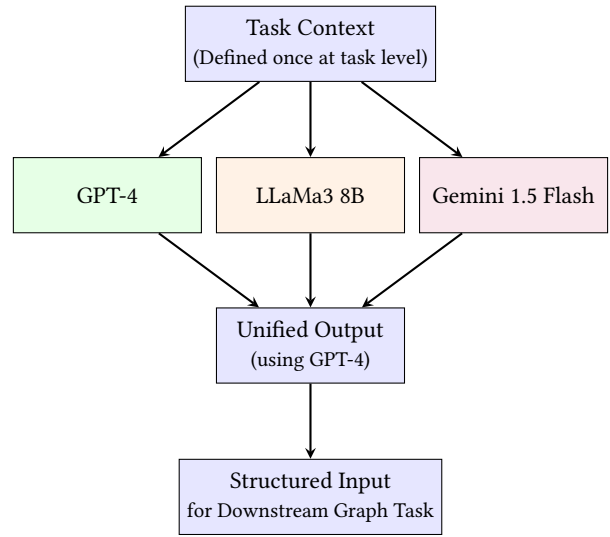
\begin{figure}[ht]
    \centering
    \begin{tikzpicture}[
        node distance=1cm and 2cm,
        box/.style={draw, rectangle, minimum width=2.5cm, minimum height=1cm, align=center, fill=blue!10},
        arrow/.style={->, thick, >=stealth}
    ]

    \node[box] (context) {Task Context \\ \small (Defined once at task level)};

    \node[box, below=of context, xshift=-2.7cm, fill=green!10] (gpt4) {GPT-4};
    \node[box, below=of context, fill=orange!10] (llama) {LLaMa3 8B};
    \node[box, below=of context, xshift=2.7cm, fill=purple!10] (gemini) {Gemini 1.5 Flash};

    \node[box, below=3cm of context] (unified) {Unified Output \\ \small (using GPT-4)};

    \node[box, below=of unified] (final) {Structured Input \\ \small for Downstream Graph Task};

    \draw[arrow] (context) -- (gpt4);
    \draw[arrow] (context) -- (llama);
    \draw[arrow] (context) -- (gemini);

    \draw[arrow] (gpt4) -- (unified);
    \draw[arrow] (llama) -- (unified);
    \draw[arrow] (gemini) -- (unified);

    \draw[arrow] (unified) -- (final);

    \end{tikzpicture}
    \caption{Overview of structured context injection for efficient graph understanding with LLMs. Different LLM outputs are consolidated into a unified input that improves downstream graph task performance. Task-specific context is defined once and then used by all questions in the task. Refer to Listings \ref{lst:orig_prompt}, \ref{lst:graph_construct} and \ref{lst:enhanced_prompt} for examples.}
    \label{fig:automation}
\end{figure}

We propose a method for structured context injection, where task-specific and graph-aware information is embedded into LLM inputs to improve performance on graph problems. Instead of using generic queries, we incorporate relevant structural details, such as node and edge relationships, tailored to the problem type (e.g., connectivity, reachability). For example, rather than asking “Is there a connection?”, the input includes the specific graph configuration and nodes of interest. Importantly, this context is defined only once at a task level and reused across related queries, ensuring efficiency. To ensure diversity and quality, we generate context using multiple LLMs (e.g., GPT-4, LLaMa3 8B, Gemini 1.5 Flash), then consolidate their outputs into a unified, high-quality input using GPT-4. This structured input formulation improves reasoning and accuracy while avoiding fine-tuning or costly multi-step querying. An overview of this process is shown in Figure~\ref{fig:automation}.

\subsection{Structured Entity-Based Graph Construction Using LLMs}

To improve LLM interpretability on graph tasks, we propose constructing graphs with meaningful entity names and relationship-aware edges. Rather than using abstract or random nouns \cite{fatemi2024talk}, we leverage well-known entities, such as characters from \textit{Game of Thrones} (GOT), whose relationships are rich, grounded, and easily interpretable. This noun-based prompting enables LLMs to draw upon their pretrained world knowledge, with edge weights capturing the nuances of relationship strength (e.g., close ties like Jon Snow and Arya Stark versus antagonistic ones like Cersei and Tyrion). Although we use GOT as a representative case, our approach can generalize to any setting with well-defined relationships, such as historical figures, real-world networks, or other narratives, making it broadly applicable for graph reasoning tasks.

To automate relationship graph generation, we use LLMs (e.g., GPT-4, Gemini, LLaMa3) following the pipeline in Figure~\ref{fig:automation}:

\begin{itemize}
  \item \textbf{Character Identification:} Prompt the model to identify major and minor characters as node candidates.
  \item \textbf{Relationship Extraction:} Elicit descriptions of known relationships (e.g., alliances, rivalries) to define graph edges.
  \item \textbf{Edge Weighting:} Infer edge weights based on interaction strength, inferred from narrative context.
  \item \textbf{Graph Structuring:} Compile the node and edge data into a formal graph for downstream tasks.
\end{itemize}

This method injects narrative context into the graph, enabling more effective reasoning on tasks like connectivity and shortest paths—without fine-tuning or manual construction (Listing \ref{lst:graph_construct}).

\begin{algorithm}[t]
\small
\SetAlgoNlRelativeSize{-1}
\caption{Approximate Subgraph Matching}
\label{alg:app_sub}
\KwIn{QueryGraph $G_q$: A graph with numeric node IDs and edges, SerialGraph $G_s$: A graph where nodes are characters and edges are relationships with weights}
\KwOut{BestSubgraph $G_{bs}$: The subgraph from $G_s$ that best matches $G_q$}

Initialize: $BestScore \gets \infty$\ , $G_{bs} \gets \text{None}$\;

\For{each Subgraph $G_{sub}$ in GenerateSubgraphs($G_s$)}{
    $Score \gets$ GraphEditDistance($G_q$, $G_{sub}$)\;
    $WgtScore \gets Score + $ CalcEdgeWeightPenalty($G_q$, $G_{sub}$)\;

    \If{$WgtScore < BestScore$}{
        $BestScore \gets WgtScore$\;
        $G_{bs} \gets G_{sub}$\;
    }
}

\KwRet{$G_{bs}$}
\end{algorithm}

\begin{algorithm}[h]
\small
\caption{Calculate Edge Weight Penalty}
\label{alg:edge_penalty}
\KwIn{QueryGraph $G_q$, Subgraph $G_{sub}$}
\KwOut{Penalty: The calculated penalty based on edge weights}

Initialize: $P \gets 0$\;

\For{each Edge $(u, v)$ in $G_q$}{
    \If{$(u, v)$ exists in $G_{sub}$}{
        $WDiff \gets |Weight(G_q, u, v) - Weight(G_{sub}, u, v)|$\;
        $P \gets P + WDiff$\;
    }
    \Else{
        $P \gets P + MaxEdgeWeight$\;
    }
}

\KwRet{$P$}
\end{algorithm}

\subsection{Subgraph Mapping for Contextual Alignment}

Prior works often replace graph node IDs with character names arbitrarily \cite{fatemi2024talk}, ignoring underlying relationships. We propose a structured mapping from numeric node IDs to a character-based graph derived from narratives like GOT. To achieve this, we employ a modified approximate subgraph matching algorithm (see Algorithm~\ref{alg:app_sub}) based on graph edit distance, with edge weight penalties reflecting relationship strengths, as described in Algorithm~\ref{alg:edge_penalty}. This approach ensures that the node-label assignments preserve the semantic and relational context inherent in the original graph.

Starting from a query graph of node IDs and undirected edges, the algorithm finds the closest matching subgraph in the character graph. The matched subgraph’s character names and relationships are then assigned to the query graph, producing a semantically rich, narrative-grounded representation. This enriched context is injected into the LLM prompt, improving reasoning accuracy by aligning raw queries with meaningful narrative structures, without requiring fine-tuning or multi-step queries (refer Listings \ref{lst:orig_prompt} and \ref{lst:enhanced_prompt}).

\begin{lstlisting}[language={},caption={Original Prompt (Graph Query)}, label={lst:orig_prompt}]
Graph: (0,4) (1,4) (2,4)
Question: Is there a path between node 0 and node 2?
\end{lstlisting}

\begin{lstlisting}[language={},caption={Graph Construction with GOT Characters and Relationship Weights}, label={lst:graph_construct}]
    #comment: The LLM's ability to recall detailed GOT facts suggests prior exposure to such data during pretraining.
    You are given an undirected weighted graph representing relationships between characters from the Game of Thrones series. 
    Each node represents a character, and each edge indicates the strength of their relationship on a scale of 1 (weak) to 5 (strong).
    
    Character-level Node Mapping:
    - 0: Jon Snow
    - 1: Arya Stark
    - 2: Sansa Stark
    - 3: Cersei Lannister
    - 4: Tyrion Lannister
    - 5: Brienne of Tarth
    - 6: Jaime Lannister
    
    Edges and Relationship Strengths:
    - (0, 4): Jon Snow and Tyrion Lannister, strong alliance (weight = 5)
    - (1, 4): Arya Stark and Tyrion Lannister, weak connection (weight = 2)
    - (2, 4): Sansa Stark and Tyrion Lannister, moderate alliance (weight = 3)
    - (3, 4): Cersei and Tyrion, hostile relation (weight = 1)
    - (2, 5): Sansa and Brienne, strong bond (weight = 5)
    - (5, 6): Brienne and Jaime, strong bond (weight = 4)
    - (3, 6): Cersei and Jaime, strong sibling bond (weight = 4)
\end{lstlisting}

\begin{lstlisting}[language={},caption={Enhanced Prompt with Narrative Context}, label={lst:enhanced_prompt}]
    You are given a weighted subgraph of Game of Thrones characters. Nodes represent characters, and edges represent their relationships. 
    Weights indicate alliance strength (1 = weak, 5 = strong).

    #comment: extract subgraph using Algorithm 1.
    Subgraph extracted from the full character graph:
    Node mapping:
    - 0: Jon Snow
    - 1: Arya Stark
    - 2: Sansa Stark
    - 4: Tyrion Lannister
    
    Relevant edges:
    - Jon Snow (0) -- Tyrion Lannister (4): weight = 5
    - Sansa Stark (2) -- Tyrion Lannister (4): weight = 3
    
    Contextual background:
    Jon Snow and Tyrion Lannister forged a strong alliance during their time at the Wall and later collaborated in Daenerys Targaryen's war council. 
    Tyrion and Sansa were once married in King's Landing, which, though politically motivated, laid a foundation for mutual respect over time.
    
    Question: Considering these relationships, is there a path between Jon Snow (node 0) and Sansa Stark (node 2) in this network?
\end{lstlisting}

\begin{table*}[t]
\small
\centering
\caption{Performance comparison across four graph reasoning tasks on Meta-LLaMA-3-8B-Instruct (LLaMA 3 8B) and Gemini 1.5 Flash. Our methods, GOT Random and GOT Subgraph, consistently achieve top accuracy across tasks. The best three scores per column are highlighted using \colorbox{top1}{green (1st)}, \colorbox{top2}{blue (2nd)}, and \colorbox{top3}{orange (3rd)} for clarity.}

\label{tab:graph_tasks}
\begin{tabular}{l|cc|cc|cc|cccccc}
\toprule
\multirow{2}{*}{Method} &
\multicolumn{2}{c|}{Connectivity} &
\multicolumn{2}{c|}{Cycle Detection} &
\multicolumn{2}{c|}{Topological Sort} &
\multicolumn{6}{c}{Shortest Path} \\
\cmidrule(lr){2-3} \cmidrule(lr){4-5} \cmidrule(lr){6-7} \cmidrule(lr){8-13}
& LLaMA & Gemini & LLaMA & Gemini & LLaMA & Gemini & \multicolumn{3}{c}{LLaMA} & \multicolumn{3}{c}{Gemini} \\
\cmidrule(lr){8-10} \cmidrule(lr){11-13}
& & & & & & & Path & Weight & Both & Path & Weight & Both \\
\midrule
Zero-shot    & 63.64\% & 67.33\% & 50.00\% & 42.67\% & 0.00\% & 12.77\% & 6.11\% & 3.06\% & 1.94\% & 30.00\% & 21.94\% & 18.06\% \\
Few-shot     & 42.90\% & 85.51\% & 49.33\% & 51.33\% & 10.00\% & 38.33\% & 31.39\% & 20.83\% & 16.67\% & 62.22\% & 59.17\% & 56.67\% \\
0-CoT        & 71.88\% & 82.95\% & 46.67\% & 44.67\% & 8.33\% & 16.11\% & 17.50\% & 13.33\% & 11.39\% & 40.00\% & 26.67\% & 23.89\% \\
CoT          & 70.17\% & 88.07\% & 49.33\% & 51.33\% & 29.72\% & 34.44\% & \cellcolor{top3}46.67\% & \cellcolor{top3}48.06\% & \cellcolor{top3}43.61\% & \cellcolor{top2}86.39\% & \cellcolor{top2}88.06\% & \cellcolor{top2}86.39\% \\
Algorithm    & 65.63\% & 85.80\% & \cellcolor{top3}53.33\% & 52.00\% & 20.00\% & 25.88\% & 38.89\% & 30.00\% & 27.22\% & 82.78\% & 83.89\% & 82.78\% \\
Instruct     & \cellcolor{top3}73.01\% & \cellcolor{top3}89.72\% & 52.67\% & \cellcolor{top3}54.00\% & \cellcolor{top3}31.83\% & \cellcolor{top3}39.73\% & 45.28\% & 44.72\% & 41.11\% & \cellcolor{top3}86.11\% & \cellcolor{top3}87.78\% & \cellcolor{top3}86.11\% \\
Noun-based   & 62.16\% & 83.49\% & 46.00\% & 49.33\% & 24.44\% & 33.27\% & 37.67\% & 39.06\% & 34.61\% & 78.50\% & 79.89\% & 78.50\% \\
\hline
\textbf{GOT Random} (Ours)   & \cellcolor{top2} 77.56\% & \cellcolor{top2} 91.19\% & \cellcolor{top2} 54.00\% & \cellcolor{top2} 71.33\% & \cellcolor{top2} 32.78\% & \cellcolor{top2} 51.94\% & \cellcolor{top2} 52.22\% & \cellcolor{top2} 55.56\% & \cellcolor{top2} 50.83\% & 83.33\% & 85.00\% & 83.33\% \\
\textbf{GOT Subgraph} (Ours) & \cellcolor{top1}80.68\% & \cellcolor{top1} 95.74\% & \cellcolor{top1} 59.33\% & \cellcolor{top1} 78.67\% & \cellcolor{top1} 34.44\% & \cellcolor{top1} 57.55\% & \cellcolor{top1} 55.83\% & \cellcolor{top1} 59.17\% & \cellcolor{top1} 54.44\% & \cellcolor{top1} 90.00\% & \cellcolor{top1} 91.39\% & \cellcolor{top1} 88.89\% \\
\bottomrule
\end{tabular}
\end{table*}

\subsection{Enhanced Prompts}
To ensure efficiency, we introduce two streamlined methods: \textbf{GOT Random} and \textbf{GOT Subgraph}. Both methods require only a \textbf{single LLM call per question}, avoiding multi-step reasoning or iterative prompting. The graph construction process is described in Listing~\ref{lst:graph_construct}, and the prompt used by both approaches is shown in Listing~\ref{lst:enhanced_prompt}. Original prompt is in Listing \ref{lst:orig_prompt}. The key difference lies in how character names are assigned to graph nodes: GOT Random assigns names randomly, without considering graph structure, while GOT Subgraph uses an approximate subgraph matching algorithm to assign names based on structural and relational similarity to the reference character graph. This allows GOT Subgraph to produce more contextually coherent and narratively grounded prompts. 

\section{Experiments}

\subsection{Experimental Setup}
We evaluate our method on four graph tasks, namely, connectivity, cycle detection, topological sorting, and shortest path, using the easy split from \cite{wang2024canNLGraph}. We test two models: Meta-LLaMa-3-8B-Instruct (open-access) and Gemini 1.5 Flash (proprietary). For LLaMa, inference is run in FP16 using PyTorch 2.0 on an NVIDIA A100 80GB GPU with 128GB RAM. All experiments use consistent settings: 512-token output length, temperature 0.001, and top-p 0.95. Subgraph comparisons in Algorithm~\ref{alg:app_sub} are capped at 25. This setup enables a controlled evaluation of prompting strategies across tasks.

\subsection{Baselines}

We compare several prompting strategies: zero-shot, few-shot, chain-of-thought (CoT), algorithm, instruction, and noun-based methods \cite{wang2024canNLGraph,fatemi2024talk}. In the noun-based, node IDs are replaced with GOT character names, following \cite{fatemi2024talk}. We exclude more complex methods like CoT with self-consistency, ToT, GoT, BoT, and MCTS \cite{yao2023tree,zhang2024accessingMCTS,besta2024graph,yang2024buffer} due to their high computational cost (multi-step querying), scalability issues, and unstable reasoning behavior. Accuracy is used as the evaluation metric, with task-specific criteria for correctness.

\subsection{Results}

Table~\ref{tab:graph_tasks} compares various prompting strategies across four graph reasoning tasks, evaluated on LLaMA 3 8B and Gemini 1.5 Flash. Our methods, \textbf{GOT Random} and \textbf{GOT Subgraph}, consistently outperform nearly all baselines, demonstrating the effectiveness of structured, narrative-grounded context injection. Our results show a clear trade-off between model size and performance. While the larger Gemini 1.5 Flash consistently outperforms the smaller LLaMA 3 8B across most tasks, the LLaMA model remains competitive, especially when enhanced with our context injection methods. This highlights that structured prompting can significantly boost performance even for smaller, more efficient models, offering a practical balance between computational cost and accuracy.

GOT Subgraph achieves the highest accuracy across multiple tasks, with \textbf{80.68\%} (LLaMA 3 8B) and \textbf{95.74\%} (Gemini 1.5 Flash) in connectivity, outperforming the best baseline, instruction-based prompting, by 6–7\%. For cycle detection, GOT Subgraph leads with \textbf{59.33\%} and \textbf{78.67\%}, surpassing the algorithm-based baseline by over 6 percentage points. In topological sorting, it also ranks highest with \textbf{34.44\%} and \textbf{57.55\%}, exceeding instruction-based baselines by 2–3\%. On the shortest path task, GOT Subgraph attains top scores across all metrics, reaching up to \textbf{90.00\%} accuracy on path prediction with Gemini 1.5 Flash, outperforming the next best method (CoT) by a clear margin. GOT Random closely follows, confirming the robustness of our narrative-based context.

Both LLaMA 3 8B and Gemini 1.5 Flash models benefit significantly from our proposed methods, clearly demonstrating the approach’s broad and practical applicability. The observed improvements stem primarily from enhanced semantic alignment through structured subgraph matching, rather than model-specific advantages, further highlighting the effectiveness of narrative-grounded context in improving diverse graph reasoning tasks.

\section{Conclusion}
We proposed a scalable and efficient framework for enhancing LLM performance on graph reasoning tasks through structured context injection. By embedding task-specific and narratively grounded information directly into prompts, our method avoids the need for fine-tuning or multi-step querying while delivering consistent performance gains. Evaluation across multiple graph tasks and model scales highlights the trade-offs between cost and accuracy, with our approach outperforming or matching more complex strategies. These results affirm structured context injection as a practical and generalizable solution for efficient graph understanding with LLMs.




\bibliographystyle{ACM-Reference-Format}
\bibliography{sample-sigconf}



\end{document}